\documentclass[conference,a4paper]{APSIPA2025}
\usepackage{xcolor}
\usepackage{amsmath}
\usepackage{graphicx}
\usepackage{multirow}
\usepackage{threeparttable}
\usepackage{booktabs}
\usepackage{hyperref}
\usepackage{tabularx}
\usepackage{fancyhdr}
\usepackage{amsfonts}
\usepackage[backend=biber,style=ieee,]{biblatex}
\usepackage{footmisc}
\usepackage{geometry}
\usepackage{fancyhdr}

\fancypagestyle{firststyle}{
  \fancyhf{}
  \fancyhead[C]{2025 Asia Pacific Signal and Information Processing Association Annual Summit and Conference (APSIPA ASC)}
}

\begin{document}

\title{REAL-WORLD MUSIC PLAGIARISM DETECTION WITH MUSIC
SEGMENT TRANSCRIPTION SYSTEM}

\author{
\authorblockN{
Seonghyeon Go
}

\authorblockA{
Mippia Inc. \\
E-mail: gsh@mippia.com}
}

\maketitle
\thispagestyle{firststyle}
\pagestyle{fancy}

\begin{abstract}
As a result of continuous advances in Music Information Retrieval (MIR) technology, generating and distributing music has become more diverse and accessible. In this context, interest in music intellectual property protection is increasing to safeguard individual music copyrights.
In this work, we propose a system for detecting music plagiarism by combining various MIR technologies. We developed a music segment transcription system that extracts musically meaningful segments from audio recordings to detect plagiarism across different musical formats. With this system, we compute similarity scores based on multiple musical features that can be evaluated through comprehensive musical analysis. Our approach demonstrated promising results in music plagiarism detection experiments, and the proposed method can be applied to real-world music scenarios. We also collected a Similar Music Pair (SMP) dataset for musical similarity research using real-world cases. The dataset are publicly available.\footnote{\href{https://github.com/Mippia/smp_dataset}{https://github.com/Mippia/smp\_dataset}}

\end{abstract}
\section{Introduction}\label{sec:introduction}
Music plagiarism is one of the most important copyright issues in society. The unauthorized copying of musical elements can have serious legal and economic consequences~\cite{yuan2023perceptual}. Contrary to the definition of a word, the commonly used word "music plagiarism" can be controversial enough even if it is not intentional by the musician. Therefore, technology for detecting plagiarism can be useful for both original composers and alleged plagiarists. With the advancement of AI music generation, creating and distributing music has become more accessible, making plagiarism detection important.

Research on defining musical similarity and detecting music plagiarism has been conducted widely~\cite{park2022music}\cite{he2021music}. However, applying these studies to real audio data faces several challenges. Most plagiarism detection research relies on MusicXML or MIDI formats, while commercial music exists as raw audio, requiring transcription. Also, many studies assume melodically similar music is plagiarized, but this differs significantly from real-world cases. And real plagiarism cases are complex~\cite{yuan2023perceptual}, potentially including vocals, varying in length, or containing brief plagiarized segments within longer tracks. A proper model needs to identify musically meaningful segments and detect plagiarism within them.

To address these issues, we propose transcribing raw audio into musical representations to organize essential musical features. Our goal is extracting musically meaningful and quantized data for plagiarism detection. Although similar ideas exist~\cite{roman2019holistic}, we focus on creating structured segment optimized for plagiarism detection by combining various music information retrieval techniques. Based on these quantized data, we explain how to detect plagiarized music using similarity metrics. Finally, we construct a Similar Music Pair (SMP) dataset containing metadata of similar music pairs with timestamps of similar segments.

\begin{figure*}[t]
    \centering
    \includegraphics[alt={ISMIR 2025 template example image}, width=0.9\textwidth]{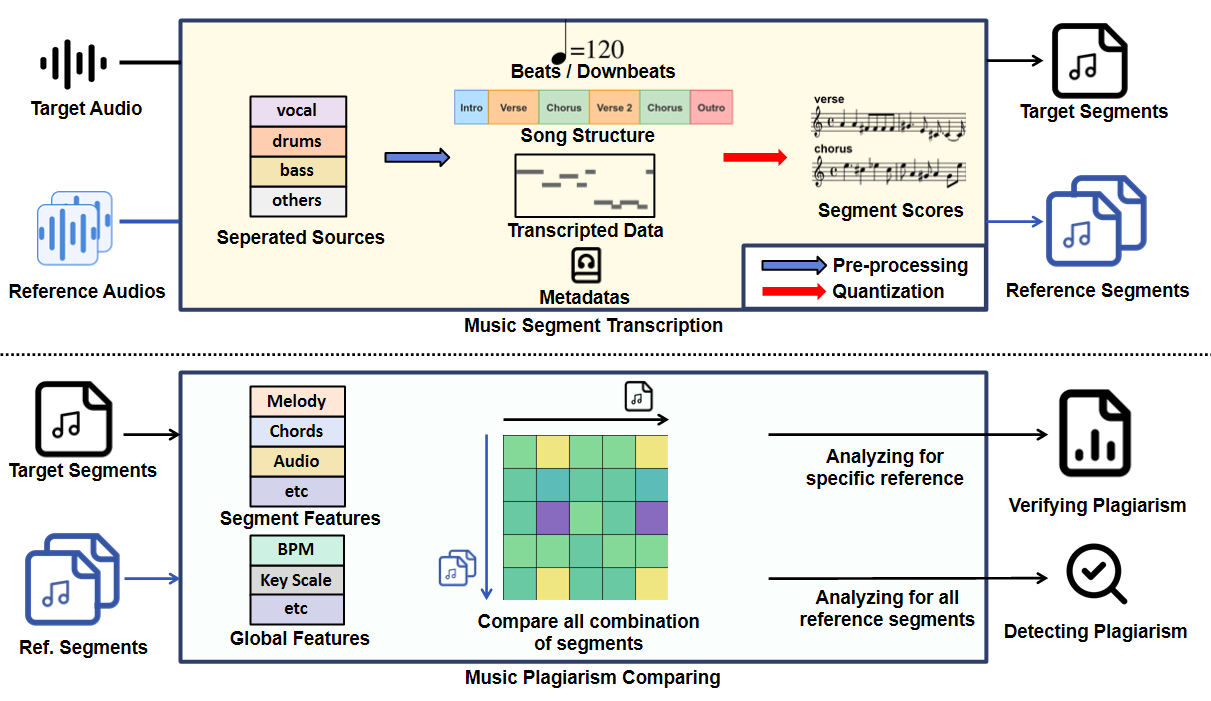}
    \caption{Overall structure of music plagiarism detection system}
    \label{fig:overall}
\end{figure*}

\section{Related Works}

\subsection{Music Transcription}
Music transcription extracts note information from raw audio, typically producing MIDI representations. This task has been studied across various genres and instruments ~\cite{donahue2021sheet} or metadata like lyrics~\cite{deng2022end}. Beyond MIDI, there is growing interest in transcribing audio into music-score-like representations~\cite{bukey2024just}. This approach allows transcription of complete musical progressions with temporal components, such as measures.

We propose segment transcription that incorporates music structure analysis and introduces metric-based self-similarity to analyze and transcribe music segments while adding musical information.

This approach allows detailed and musically meaningful results beyond calculating similarity. For instance, we could pinpoint that music A's first chorus segment (e.g., 00:35–00:43, 16th–19th bar) corresponds to music B's second chorus segment (e.g., 01:42–01:51, 44th–47th bar). We define this musical unit as a 'Segment'. This methodology enables detection of similar or plagiarized segments across larger datasets.

To perform music segment transcription, we combined MIR technologies including music source separation, beat-tracking, chord recognition to obtain necessary metadata and construct better structural representations for each segment.

\subsection{Music Plagiarism Detection}
Research on music plagiarism analysis employs various methodologies, including CNN-based approaches~\cite{park2022music}, bipartite graph-based methods~\cite{he2021music}, NLP-based methods using tokenization~\cite{malandrino2022adaptive}, and audio fingerprinting-based methods~\cite{borkar2021music}. However, most approaches focus on MIDI or MusicXML data, with limited methodologies using raw audio data~\cite{de2015plagiarism}. But in applying plagiarism detection to real-world scenarios, using raw audio data is essential.

Cover Song Identification (CSI) can be considered a similar task to plagiarism detection, as it involves retrieving cover versions of a query song from a music dataset. Classically, there is a melody MIDI-based methodology~\cite{marolt2006mid} or a methodology for interpreting and comparing a sound source as a sequence~\cite{serra2009cross}. Recently, the trend of studying CSI based on deep learning is increasing. Various methods using CNN have been proposed~\cite{yu2020learning,du2022bytecover2,du2023bytecover3}. In addition to this, models using the conformer structure with music chunk~\cite{liu2023coverhunter} have achieved SOTA performance. However, CSI methods typically perform song-to-song comparisons to determine overall similarity between entire musical works. In contrast, plagiarism detection requires more fine-grained analysis to identify specific passages where copying occurs, necessitating segment-level comparisons that can pinpoint exactly which parts of songs are similar and potentially plagiarized.

A key challenge in plagiarism detection is defining MIDI or audio similarity. Some approaches use shape similarity~\cite{urbano2011melodic}, while others tokenize notes and apply sequence-based deep learning models~\cite{karsdorp2019learning}. Other approaches train models to learn features similar to plagiarism cases, using embedding distance as metrics~\cite{park2022music}. We propose a segment-based methodology that decomposes songs into meaningful musical segments and performs individual comparisons between segments. This approach enables detection of partial plagiarism where only specific sections are borrowed and provides precise localization of similar passages. We combine various metrics after music segment transcription, focusing on shape-based similarities and musical feature similarities.

\begin{table*}[t]
\vspace{0.5em}
\centering
\small
\caption{Necessary information from segments in music plagiarism detection scenarios}
\begin{tabular}{|p{7.5cm}|p{7.5cm}|}
\hline
\multicolumn{2}{|p{15cm}|}{\textbf{Scenario:} We have music A, and a dataset including music B. We want to determine that "The chorus vocal melody in music A from 30-38 seconds is plagiarized from music B's chorus vocal melody at 40-47 seconds."} \\
\hline
\textbf{Critical Questions} & \textbf{Necessary Information} \\
\hline
1. What is the basis for selecting \textit{30s} as the segment starting point in music A? How about B?& 
Music structure analysis, downbeat, BPM \\
\hline
2. What is the basis for selecting \textit{38s} as the segment ending point in music A? How about B?& 
Downbeat, rhythm, BPM \\
\hline
3. How is the length discrepancy (8s vs 7s) addressed when comparing segments? & 
Quantized note information \\
\hline
4. What evidence supports that the \textit{chorus} aspect is plagiarized rather than other musical structures? & Music structure information for each segment \\
\hline
5. What evidence supports that specifically the \textit{vocal} component is plagiarized? & 
Instrument-specific transcription \& similarity \\
\hline
6. What evidence supports that the \textit{melody} aspect is plagiarized rather than other musical elements? & Similarity for each musical element \\
\hline
7. What evidence supports that \textit{music B} is the source rather than other pieces? &
Similarities from all reference segments \\
\hline
\end{tabular}

\label{tab:plagiarism_scenario}
\end{table*}

\section{Proposed System}

The overall structure is illustrated in Figure \ref{fig:overall}. The target audio is processed through the music segment transcription system. After preprocessing and quantization, the audio is analyzed in distinct segments. Each segment contains musical information such as melody, chords, instruments, etc. These segments are then compared for similarity with other preprocessed reference segments, enabling plagiarism detection. In the following sections, we provide a detailed explanation of how the music segment transcription process is carried out and how the similarity computation is performed.

\subsection{Music Segment Transcription}
We first preprocess music to find necessary components for quantization, and then quantize music into segments. Necessary information from segments is described in Table \ref{tab:plagiarism_scenario}, with a music plagiarism detection scenario. The preprocessing required for this task includes source separation, transcription results of each separated source, chord recognition results, music structure information, tempo, downbeat, time signature, and optionally, lyric information or various metadata of the music.

To transcribe music segments, we first identify the bpm and downbeats of the waveform $w$ and quantized to a arithmetic sequence, denoted as $db_{w}=\{dbt_1, dbt_2, ..., dbt_n\}\in\mathbb{R}^{n}$. We assume that the segment start points align with $db_{w}$. We then identify the music structure to locate the points of structural change. These music structure boundaries are represented as $ss_{w}=\{ss_1, ss_2, ..., ss_m\}\in\mathbb{R}^{m}$, which is quantized to be a subset of $db_{w}$. Since the length of each music structure is not always consistent (e.g., 4 bars or 8 bars), and even segments with the same structural role may exhibit distinct patterns (e.g., verse1 and verse2), we perform additional segment clustering to properly identify and group these varying musical elements.

We use a self-similarity-based clustering algorithm to find each start point of segments, identifying and organizing recurring core patterns that frequently appear in music. A similarity matrix is constructed by computing the similarity between all available segment start points from $db_w$. Then, we apply hierarchical clustering using Ward's method~\cite{ward1963hierarchical}.
Ward's method minimizes the variance within clusters by recursively merging the pair of clusters that leads to the minimum increase in total within-cluster variance. The number of clusters is determined proportionally to the size of the distance matrix, with an appropriate threshold. After clustering, the most frequently occurring patterns in music are treated as the most significant. Note that the "distance" can be defined in any methods. Used distance between each segment is discussed in Section 3.2, which has an inverse relationship with similarity.

For each boundary identified in the music structure, we extract a potential starting point and refine these initial points by ensuring they align with musical phrases, with more segments aligned in clusters that have high priority, typically occurring at 4-bar intervals. This process yields a set of refined starting points $SP_w = \{sp_1, sp_2, ..., sp_k\}$ that accurately delineate the beginning of each significant musical segment, ensuring that our segmentation respects the inherent structural organization of the music. This approach tends to create compact segments and is effective for our segment analysis as it groups similar musical patterns while maintaining boundaries from music structure information. 
These starting points mark where each music segment begins and help us quantize the music's structure. For each starting point, we define a segment as the interval spanning a fixed length, typically 4 bars for this work.

\begin{table*}[t]
\centering
\caption{Sample entries from our Similar Music Pair dataset}
\begin{tabular}{|l|l|l|l|l|l|}
\hline
\textbf{Original Title} & \textbf{Comparison Title} & \textbf{Relation} & \textbf{Original Time} & \textbf{Comparison Time} & \textbf{Pair \#} \\
\hline
Electric & Electric (remix)  & Remake & [8, 16, 70, 78] & [7, 15, 82, 90] & 21 \\
\hline
Shiki no uta & Bul-ggot & Plagiarism Case & [73, 82, 134, ...] & [85, 93, 136, ...] & 29\\
\hline
Volevo un gatto nero & Black Cat Nero & Remake & [25, 33, 59,...] & [68, 76, ...] & 31\\
\hline
No scrubs & Shape of you & Plagiarism Case & [31, 41, 72, ...] & [15, 96] & 64 \\
\hline
\end{tabular}

\label{tab:dataset_sample}
\end{table*}
We obtain transcription data for each instrument, $N_{(i,\mathcal{I})} = (p_{(i,\mathcal{I})} , t_{(i,\mathcal{I})} , d_{(i,\mathcal{I})}, v_{(i,\mathcal{I})})$ where $p$ denotes pitch, $t$ denotes onset time, $d$ denotes duration, $v$ denotes velocity in MIDI, and $\mathcal{I}$ is the set of instrument types. We obtain the start time and duration in seconds. With BPM and downbeat information, we can determine which $sp_i$ these notes correspond to and quantize their positions. Let $QN_{(i,\mathcal{I})} = (p_{(i,\mathcal{I})}, b_{(i,\mathcal{I})}, pos_{(i,\mathcal{I})}, qd_{(i,\mathcal{I})}, v_{(i,\mathcal{I})}) \in \mathbb{Z}^5$ represent this quantized note information, where $b$ is bar number in music score, $pos$ is quantized onset in corresponding bar, and $qd$ is quantized duration. For example, we can state "The E5 vocal note exists as a quarter note on the second beat of 14th bar with a velocity of 100, and bars 12--15 comprise the verse 2 segment."
Through this process, we analyze and organize the given musical data similar to sheet music format. This structured data is directly used in similarity comparison tasks.

\subsection{Music Plagiarism Detection}\label{section 3.2}

Our similarity calculation incorporates multiple musical aspects. Each aspect is structured in a algorithmic manner, allowing consideration of melody and chords based on cases found in various music plagiarism scenarios~\cite{yuan2023perceptual}:
\begin{itemize}
\item \textbf{Pattern Similarity} $p$: The chromagram-based intersection similarity.
\item \textbf{Musical Complexity} $m$: Count of used pitches, weighted to $p$ to avoid overly simple similarity cases, such as rap.  
\item \textbf{Rhythmic Correlation} $r$: Jaccard similarity with quantized onset timing.
\item \textbf{BPM Difference Ratio} $b$: Linear scaling of tempo relationship.
\item \textbf{Chord Similarity} $c$:
\begin{itemize}
\item \textbf{Roman numeral similarity} $R_n$: Functional harmony-based comparison~\cite{weber1832versuch}
\item \textbf{Chord quality similarity} $Q$: Major/minor and seventh chord relationships
\end{itemize}
\end{itemize}
\begin{equation}
c = w_R \cdot R_n + w_Q \cdot Q
\end{equation}
\begin{equation}\label{similarity}
\text{similarity} = (\alpha + m \times p) \times (\beta \cdot \max(r, p)) \times b^{\delta} + \gamma \cdot c
\end{equation}

Equation~\ref{similarity} is one example of our similarity metric used in experiments. The parameters $\alpha$, $\beta$, $\gamma$, $\delta$ and weights $W_R$, $W_Q$ balance the contribution of each musical component.

\section{Datasets}
For our experiments, we compiled a SMP dataset, which is a comprehensive dataset of music piece pairs for plagiarism detection evaluation. The SMP dataset contains 70 pairs of original and comparison music pieces, each with relevant metadatas and segment time where similar part starts. 
Table \ref{tab:dataset_sample} presents a sample of the SMP dataset. The SMP dataset was carefully curated to include a diverse range of music genres, release periods, and similarity types. It encompasses well-known plagiarism cases, legally disputed works, pieces with acknowledged influence, and even some pairs with coincidental similarities. This diversity allows for a comprehensive evaluation of our plagiarism detection approach across various scenarios encountered in real-world music copyright dispute.
Additionally, we used the Covers80 dataset for experiments, which consists of 80 groups of cover songs, with each group containing multiple versions of the same original song performed by different artists.
\section{Experiments}

\subsection{Experimental Setting}
Given the availability of various MIR models, we were able to implement a comprehensive system for the music segment transcription process. Specifically, we used Demucs~\cite{rouard2023hybrid} for source separation, the all-in-one model~\cite{kim2023all} for structural analysis, Beat-Transformer~\cite{zhao2022beat} for downbeat tracking, AST\cite{wang2021preparation} for vocal transcription, SheetSage~\cite{donahue2022melody} for melody transcription, and Harmony Transformer~\cite{chen2019harmony} for chord transcription.

To evaluate plagiarism music detection, we conducted experiments using the SMP dataset and covers80 dataset. We calculated the average ranking and accuracy to determine how plagiarized music typically ranks. We conducted this experiment with parameter settings to compare each similarity metric's characteristics. With default setting ($\alpha=1, \beta=1, \gamma=1, \delta=0.5, w_R = 0.85, w_Q = 0.15$) for all experiments, we tested chord only ($\beta=0$), MIDI only ($\gamma=0$), rhythm only (similarity$=r$), and pattern only (similarity$=p$) cases for plagiarism music detection.
\subsection{Evaluation Methodology}

We evaluate our plagiarism detection system using two complementary metrics. 

\textbf{Segment-level Evaluation:} To further validate our segment-based approach, we conducted plagiarism segment detection experiments using the Covers80 dataset. Unlike traditional cover song identification methods that perform song-to-song comparisons, our approach conducts segment-level comparisons, enabling more fine-grained analysis of musical similarities within songs and detection of potential plagiarism at the segment level.

We employ a retrieval-based evaluation where query segments are matched against a database of source segments. Performance is measured using Precision@K, which calculates the proportion of correct matches within the top-K retrieved segments. A correct match is defined as a segment pair from the same cover song group, indicating a potential plagiarism relationship.

\textbf{Song-level Evaluation:} Since there are multiple segment pair similarities in one track pair, we computed plagiarism rates by summing the 20 highest similarity scores for each pair and ranking them in descending order. Following established practices in music similarity research~\cite{he2021music}, we use three primary metrics: \textit{Top Average Index}, representing the average rank of the correct plagiarized song in our retrieval results, with a maximum penalty for undetected cases; \textit{Top-1 Accuracy}, measuring the proportion of queries where the correct match appears as the highest-ranked result; and \textit{Top-5 Accuracy}, providing a more lenient evaluation by considering whether the correct match appears within the top-5 results.

\section{Results}

For song-level detection, we aggregate segment-level similarities using a weighted scoring approach. For each query song, we extract multiple segments and retrieve the top-5 most similar segments from our database for each query segment. The final song similarity score is computed as a weighted combination of these segment-level similarities, where weights are determined by the individual segment similarity scores.

\subsection{Plagiarism Segment Detection Results}

The results in Table~\ref{tab:plagiarism_segment_results} demonstrate the effectiveness of our segment-based approach for plagiarism detection. Precision at top-100 indicates that our similarity metric successfully identifies plagiarism relationships at the segment level. 

The segment-based approach offers several advantages over traditional whole-song comparison methods:\\ (1) This provides more detailed analysis by identifying specific musical passages that contribute to plagiarism relationships. \\(2) It enables partial matching where only certain segments of songs are similar. \\(3) This offers better interpretability by highlighting which musical elements drive the similarity scores, making it particularly valuable for plagiarism detection applications.

\begin{table}[t]
\centering
\caption{Plagiarism segment detection results}
\begin{tabularx}{\columnwidth}{l|>{\centering\arraybackslash}X|>{\centering\arraybackslash}X}
\hline
\textbf{Evaluation Metric} & \textbf{Top-100} & \textbf{Top-1000} \\
\hline
Precision@K & 98.00\% & 51.80\% \\
Correct Retrievals & 98/100 & 518/1000 \\
Metric Range & [73.22, 98.63] & [58.24, 98.63] \\
\hline
\end{tabularx}
\label{tab:plagiarism_segment_results}
\end{table}

\begin{table}[t]
\centering
\caption{Examples of segment-level detection results from top-100 retrievals}
\begin{tabular}{l|l}

\hline
\multicolumn{2}{c}{\textbf{Correct Matches}} \\
\hline
\textbf{Example 1} & Score: 98.63 \\
Query & Blue Collar Man (Styx) at 118.1s \\
Source & Blue Collar Man (REO Speedwagon) at 143.6s \\
\hline
\textbf{Example 2} & Score: 95.08 \\
Query & September Gurls (Big Star) at 9.2s \\
Source & September Gurls (Bangles) at 8.0s \\
\hline
\hline
\multicolumn{2}{c}{\textbf{Incorrect Matches}} \\
\hline
\textbf{Example 3} & Score: 74.84 \\
Query & Gold Dust Woman (Sheryl Crow) at 184.2s \\
Source & Tomorrow Never Knows (Beatles) at 110.2s \\
\hline
\textbf{Example 4} & Score: 73.41 \\
Query & Hush (Milli Vanilli) at 157.4s \\
Source & Night Time Is The Right Time (Aretha Franklin) at 95.8s \\
\hline
\end{tabular}

\label{tab:segment_examples}
\end{table}

\subsection{Plagiarism Music Detection Results}

\begin{table}[b]
\centering
\caption{Plagiarism detection results with similarity conditions}
\begin{tabularx}{\columnwidth}{l|>{\centering\arraybackslash}X|>{\centering\arraybackslash}X|>{\centering\arraybackslash}X}
\hline
\textbf{Conditions} & \textbf{Avg. Index} & \textbf{Top-1 Acc.} & \textbf{Top-5 Acc.} \\
\hline
Pattern only & 13.23 & 0.2357 & 0.3143 \\
Rhythm only & 13.29 & 0.1429 & 0.2786 \\
Chord only & 12.73 & 0.1500 & 0.2929 \\
MIDI only & 11.16 & 0.2429 & 0.4071 \\
\hline
\textbf{All (SMP)} & \textbf{7.31} & \textbf{0.3786} & \textbf{0.6286} \\
\textbf{All (Covers80)} & \textbf{13.46} & \textbf{0.475} & \textbf{0.575} \\
\hline
\end{tabularx}
\label{tab:plagiarism_test}
\end{table}
The experimental results using segmentation data are presented in Table~\ref{tab:plagiarism_test}. Results demonstrate that our system can identify plagiarism pairs using only audio data. Furthermore, they validate that incorporating various musical knowledge approaches enhances the detection of musical similarities. 

Nevertheless, at the current stage, these metrics may not be considered sufficiently reliable, especially plagiarism music detection task. We present several observations regarding failure cases below:

\textbf{Lack of end-to-end segmentation model.} In this study, we performed the segmentation task by combining existing MIR systems. The integration of these systems introduces instability, as each component creates bottlenecks.

\textbf{Instability of segment similarity.} Segment musical similarity is a challenging concept to perfect at present. We expect this can be improved through future research, using both transparent and learned approaches.

\textbf{Bridging segment-level and music unit-level metrics.} The connection between segment-level similarity metrics and music unit metrics (such as cover song identification) requires further investigation. Current approaches may miss important relationships between local musical patterns and broader segment similarities, potentially affecting the overall detection accuracy.

\subsection{Ablation Studies}

For the SMP dataset, since the starting points of actually similar segments are provided, we can construct a Siamese network\cite{koch2015siamese} for similarity metrics by performing segment transcription. We conducted a simple siamese network by obtaining embeddings from MERT\cite{li2023mert} and Music2Vec\cite{li2022map}. We perform music detection task with models, with 55 training pairs and 15 test pairs. 

\begin{table}[t]
\centering
\caption{Performance with real plagiarism pairs from SMP dataset}
\begin{tabularx}{\columnwidth}{l|>{\centering\arraybackslash}X|>{\centering\arraybackslash}X|>{\centering\arraybackslash}X|>{\centering\arraybackslash}X}
\hline
\textbf{Models} & \textbf{Avg. Index} & \textbf{Top-1 Acc.} & \textbf{Top-5 Acc.} & \textbf{Loss} \\
\hline
MERT & 4.83 & 0.333 & 0.833 & 0.0639 \\
Music2Vec & 5.00 & 0.333 & 0.667 & 0.0432 \\
\hline
\end{tabularx}
\label{tab:Pair Training}
\end{table}
 The results are presented in Table ~\ref{tab:Pair Training}. This is showing that this approach has sufficient research potential, but they also indicate that larger-scale data and better methodologies are still remain as future work.

\section{Future Work and Conclusion}

In this paper, we proposed a music plagiarism detection system that effectively handles real-world audio data through music segment transcription. By combining various MIR technologies and introducing a segment-based analysis framework, our approach demonstrates promising performance while maintaining practical applicability to commercial music. This shows that segment-based transcription can effectively bridge the gap between theoretical plagiarism detection research and real-world applications.

Several research directions can enhance this system further. Each MIR component (source separation, transcription, beat tracking) could be improved for more stable segment transcription. The development of an end-to-end approach for segment transcription represents a promising direction that could improve both efficiency and accuracy. The structured segment data format could be extended to other MIR tasks beyond plagiarism detection, including cover song detection and music generation. Future research on deep learning architectures trained on segment-level data could capture more detailed relationships between musical pieces, strengthening the system's capability to handle real-world audio data and contributing to music copyright protection and analysis.

\printbibliography

\end{document}